\documentclass{article}

\usepackage{PRIMEarxiv}
\usepackage[utf8]{inputenc}
\usepackage[T1]{fontenc}
\usepackage{hyperref}
\usepackage{url}
\usepackage{booktabs}
\usepackage{amsfonts}
\usepackage{nicefrac}
\usepackage{microtype}
\usepackage{lipsum}
\usepackage{fancyhdr}
\usepackage{graphicx}
\usepackage{listings}
\graphicspath{{media/}}
\usepackage{caption}
\usepackage{subcaption}
\usepackage{amsmath}
\usepackage{float}

\title{Applying Bayesian Ridge Regression AI Modeling in Virus Severity Prediction

}

\author{
  Jai Pal \\
  Independent Researcher \\
  \texttt{jaipal9621@gmail.com}
  \and
  Bryan Hong \\
  Independent Researcher \\
  \texttt{bzhtexas@gmail.com}
}

\begin{document}
\maketitle

\begin{abstract}
Artificial intelligence (AI) is a powerful tool for reshaping healthcare systems. In healthcare, AI is invaluable for its capacity to manage vast amounts of data, which can lead to more accurate and speedy diagnoses, ultimately easing the workload on healthcare professionals. As a result, AI has proven itself to be a power tool across various industries, simplifying complex tasks and pattern recognition that would otherwise be overwhelming for humans or traditional computer algorithms. In this paper, we review the strengths and weaknesses of Bayesian Ridge Regression, an AI model that can be used to bring cutting edge virus analysis to healthcare professionals around the world. The model's accuracy assessment revealed promising results, with room for improvement primarily related to data organization. In addition, the severity index serves as a valuable tool to gain a broad overview of patient care needs, aligning with healthcare professionals' preference for broader categorizations.
\end{abstract}

\section{Introduction}
Artificial intelligence, or AI, has proven itself to be a powerful tool across various
industries, simplifying complex tasks and pattern recognition that would otherwise be
overwhelming for humans or traditional computer algorithms. Its versatility is evident in
its ability to transform operations in many fields, and healthcare is no exception. In
healthcare, AI is invaluable for its capacity to manage vast amounts of data, which can
lead to more accurate and speedy diagnoses, ultimately easing the workload on healthcare
professionals.

The utility of AI spans far and wide, from optimizing supply chains to
revolutionizing customer service and financial forecasting. However, when it comes to
healthcare, the focus shifts to its incredible potential to handle the immense volumes of
medical data we encounter daily.

In the healthcare sector, data-driven decisions are crucial. Precise and timely
diagnoses and prognoses are paramount, and AI plays a pivotal role in achieving these
goals. It can compile and analyze millions of data points, creating comprehensive models
that assist in making medical assessments. This becomes particularly important during
critical times, such as the peak of the COVID-19 pandemic.
Throughout the pandemic, healthcare workers faced an unprecedented workload,
strained resources, and a dire need for rapid and accurate decision-making. In such
circumstances, AI modeling became a lifeline. AI tools were employed to analyze patient
data, predict disease progression, and optimize the allocation of resources. These
applications not only saved time but also helped save lives when healthcare systems were
pushed to their limits.

Furthermore, healthcare research and analysis inherently involve vast amounts of
data. This encompasses patient records, genetic information, clinical trials, and medical
imaging, creating a need for a sophisticated approach. While traditional computer
algorithms can handle large datasets, they may struggle to adapt to changing data trends
and patterns.

AI systems excel in this regard. They possess the capability to continuously learn
and adapt as new data becomes available, making them ideal for the dynamic nature of
healthcare research. Whether it's identifying rare genetic mutations linked to diseases or
predicting the outcomes of innovative treatments, AI's ability to navigate extensive
datasets and discern nuanced patterns is unparalleled.

Integrating AI into healthcare isn't just a technological advancement; it represents
a transformative shift in how we approach medical diagnoses, treatments, and research.
By harnessing AI's capacity to process and interpret vast datasets, healthcare
professionals can make more precise decisions, ultimately leading to improved patient
outcomes and a more efficient healthcare system. As AI continues to evolve, its role in
reshaping healthcare is poised to grow, benefiting patients and healthcare providers alike.

\section{Artificial Intelligence Model Selection Process}
In the context of this research, it's essential to delve further into the specifics of the
modeling approach, particularly the utilization of Bayesian Ridge Regression. The
research centered on a virion counting system that hinged on the use of plasmonic gold
nanoparticles to individually measure the virion count in samples. This method was
developed with a focus on the respiratory syncytial virus (RSV) as the target disease for
the samples. RSV was chosen because it predominantly affects infants, who, unlike
adults, typically do not have other complicating health issues or significant
predispositions.

When studying the development of diseases, certain biological truths tend to hold
across a majority of cases. While numerous parameters and factors contribute to disease
prognosis, in the case of infants, two critical factors are weight and age. These variables
are pivotal because they play interconnected roles in shaping an infant's immunity. In
general, as an infant's weight increases, its immunity also strengthens in a direct
correlation. This relationship means that even if two infants have the same virion count of
the disease, the one with a higher weight will typically experience a milder form of the
illness. However, it's crucial to note that if an infant's weight falls to an extreme, either
too high or too low, their immunity can be compromised.

The second key relationship involves age. As an infant ages, their immunity tends
to increase in a more or less linear and straightforward manner. This age-related
improvement in immunity is a vital consideration when selecting an AI model for this
type of research. Acknowledging these underlying biological principles can be
instrumental in reducing data requirements and enhancing the overall model fitting
process.

Given the unique situation and the presence of these well-defined biological laws,
Bayesian Ridge Regression was chosen as the AI model of choice. Bayesian Ridge
Regression stands out for its adaptability and ability to hyper-parameterize. This
adaptability is particularly relevant in cases like this, where the relationships between
weight, age, virion count, and disease severity are complex and may not follow
traditional linear patterns. Bayesian Ridge Regression excels in capturing and quantifying
these intricate relationships by adapting its model parameters to fit the data more
effectively.

In essence, Bayesian Ridge Regression was the ideal choice for this research due
to its ability to navigate the nuanced interplay of biological factors and its capacity to
incorporate domain-specific knowledge into the modeling process. By leveraging this AI
model, the research aimed not only to provide more accurate disease prognoses but also
to gain a deeper understanding of the intricate connections between infant health,
immunity, and disease severity, ultimately contributing to improved healthcare outcomes
for infants affected by diseases like RSV.

\section{Data Collection and Variance}
To generate testable data for the research, it was essential to define a target
variable that could be used to assess the severity of a given condition. In this particular
case, three input variables were considered: weight, age, and virion count. When these
input variables are provided, a severity index is calculated by the system, and this index
serves as a measure of the condition's severity. The code responsible for calculating this
severity index is outlined in Figure 1.

\begin{lstlisting}[language=Python, caption={Severity index calculation}, label={lst:python}]
for i in range(1000000):
age = random.randint(0, 24)
virionCount = random.randint(1, int(1e10))
gender = random.randint(0, 1)
if gender == 0:
weight = random.uniform(0, mhighWeightList[age] +mlowWeightList[age])
acceptableWeight = (mhighWeightList[age] + mlowWeightList[age]) / 2
else:
weight = random.uniform(0, fhighWeightList[age] + flowWeightList[age])
acceptableWeight = (fhighWeightList[age] + flowWeightList[age]) / 2
severity = (1 - age / 24) * (virionCount) + abs((acceptableWeight - weight)
/ acceptableWeight)*((virionCount) ** 2)
# Add 0.01% variance to the severity by multiplying it with a random factor
variance_factor = 1 + random.uniform(-0.0001, 0.0001)
severity_with_variance = severity * variance_factor
\end{lstlisting}

The calculation of the severity index involves the creation of two coefficients: an
age coefficient and a weight coefficient. These coefficients are then multiplied by the
virion count and the square of the virion count, respectively. It's worth noting that weight
is emphasized as the dominant indicator of an infant's health, as opposed to age. This
emphasis is reflected in the attribution of weight with the square of the virion count,
while age is paired with the regular virion count. The rationale behind this choice is that,
in general, weight has a more significant impact on an infant's health compared to age. As
age increases, the coefficient associated with it decreases, effectively limiting the growth
of the severity curve. Conversely, as an infant's weight deviates further from a previously
defined acceptable weight range (as shown in Figure 2 and Figure 3), the coefficient
corresponding to weight increases, which in turn fuels the growth of the severity curve.
\begin{lstlisting}[language=Python, caption={Weight lists}, label={lst:python}]
mhighWeightList = \[3.9, 5.1, 6.3, 7.2, 7.9, 8.4, 8.9, 9.3, 9.6, 10.0, 10.3, 10.5,
10.8, 11.1, 11.3, 11.6, 11.8, 12.0, 12.3, 12.5, 12.7, 13.0, 13.2, 13.4, 13.7]
mlowWeightList = [2.9, 3.9, 4.9, 5.6, 6.2, 6.7, 7.1, 7.4, 7.7, 7.9, 8.2, 8.4, 8.6, 
8.8, 9.0, 9.2, 9.4, 9.6, 9.7, 9.9, 10.1, 10.3, 10.5, 10.6, 10.8]
fhighWeightList = [3.7, 4.8, 5.9, 6.7, 7.3, 7.8, 8.3, 8.7, 9.0, 9.3, 9.6, 9.9, 10.2,
10.4, 10.7, 10.9, 11.2, 11.4, 11.6, 11.9, 12.1, 12.4, 12.6, 12.8, 13.1]
flowWeightList = [2.9, 3.6, 4.5, 5.1, 5.6, 6.1, 6.4, 6.7, 7.0, 7.3, 7.5, 7.7, 7.9,
8.1, 8.3, 8.5, 8.7, 8.8, 9.0, 9.2, 9.4, 9.6, 9.8, 9.9, 10.1]
\end{lstlisting}

\begin{lstlisting}[language=Python, caption={Acceptable weight calculation}, label={lst:python}]
if gender == 0:
weight = random.uniform(0, mhighWeightList[age] + mlowWeightList[age])
acceptableWeight = (mhighWeightList[age] + mlowWeightList[age]) / 2
else:
weight = random.uniform(0, fhighWeightList[age] + flowWeightList[age])
acceptableWeight = (fhighWeightList[age] + flowWeightList[age]) / 2
\end{lstlisting}

It is important to note that the calculation for the severity index is not constant
amongst all diseases and will change thoroughly depending on research and clinical
observation. The index calculation used in this model is purely for training purposes and
as a proof of concept. In a real world scenario, the severity index will likely illustrate a
starkly different curve.

In the process of training the AI model, a substantial dataset comprising one
million combinations of weight, age, virion count, and their respective severity indices
was generated. This dataset served as the foundation for training the model. To ensure
robust model evaluation, the dataset was divided into two parts: eighty percent of the data
was allocated for training, while the remaining twenty percent was set aside for testing
and validation purposes. This division allowed for a thorough assessment of the model's
performance on unseen data.

To simulate real-world scenarios and account for potential discrepancies, an
additional dataset was created. This dataset introduced a variance of 0.01 percent, aiming
to replicate the inherent uncertainties and variations encountered in practical healthcare
settings. This second dataset was tested separately from the precise dataset, enabling the
evaluation of the model's robustness and its ability to handle data with real-world noise.

In summary, the process of generating testable data involved the definition of a
severity index based on weight, age, and virion count. The intricate calculations behind
this severity index incorporated coefficients that accounted for the influence of weight
and age on an infant's health. The dataset used for model training and testing was
carefully constructed to ensure comprehensive evaluation, including the incorporation of
real-world variations to enhance the model's applicability in practical healthcare
scenarios.

\section{Addressing Inaccuracies}
One of the primary challenges encountered when organizing the data for analysis
lies in the inherent inaccuracy of the severity index itself. This inaccuracy is primarily
attributed to the dynamic and volatile nature of the age and weight coefficients within the
calculation process. The interplay between these coefficients and the main variable,
virion count, can pose a formidable challenge for the AI model, making it difficult to
discern clear and consistent patterns.

To mitigate this challenge and enhance the model's effectiveness, one potential
workaround involves simplifying the complexity of the data by eliminating one of the
parameters. This approach could entail developing separate models for distinct subsets of
the data. For instance, a logical application of this approach might be to create individual
models tailored to specific age groups, such as infants aged 0-24 months, or for infants
falling within a specific weight range that deviates consistently from the acceptable
values. By doing so, the complexity of the patterns within the data would be dramatically
reduced, resulting in a more manageable dataset. This, in turn, would alleviate some of
the computational and modeling challenges associated with the intricate relationships
between age, weight, and virion count.

However, it's crucial to acknowledge that while simplifying the dataset in this
manner may enhance the model's ability to identify patterns, it also comes with
trade-offs. For instance, by segmenting the data into smaller subsets based on age or
weight, the model may become less adaptable to broader variations and trends within the
entire population. Additionally, this approach may require the development of multiple
specialized models, each tailored to a specific demographic, which can increase the
complexity of model management and maintenance.

Furthermore, it's important to consider that when dealing with rarer diseases and
viruses, quantifying their severity and accommodating them into comprehensive severity
predictions can be particularly challenging. This difficulty arises from the limited
availability of clinical data and information pertaining to these rare conditions. As a result, the AI model may struggle to accurately assess the severity of such cases due to
the scarcity of relevant training data.

In conclusion, addressing the inaccuracy associated with the severity index in data
organization is a complex task. While simplifying the dataset by segmenting it into
subsets based on age or weight may reduce complexity and data volume requirements, it
also introduces challenges related to adaptability and model management. Additionally,
for rare diseases, the scarcity of clinical data poses a unique obstacle in accurately
predicting severity. Finding the right balance between data simplification and preserving
the model's ability to capture broader trends is key to developing effective AI models for
healthcare applications.

\section{Results and Error Analysis}
To assess the accuracy of the developed model, a rigorous evaluation process was
employed. Initially, 20\% of the original dataset was deliberately set aside and partitioned
into a dedicated testing set. The remaining 80\% of the data was used for training the
model. Subsequently, the trained model faced a formidable challenge: it was tasked with
determining severity indexes for a substantial set of 200,000 combinations of weight, age,
and virion count.

\begin{lstlisting}[language=Python, caption={Model creation}, label={lst:python}]
# Load data from CSV files
X_data = pd.read_csv('x_data.csv')
# Assuming X_data.csv contains 'Weight,' 'Age,' 'Virion Count,' and 'Gender'
Y_data_variance = pd.read_csv('y_data_variance.csv')
# Assuming Y_data_variance.csv contains 'Severity' with variance
Y_data_precise = pd.read_csv('y_data_precise.csv')
# Assuming Y_data_precise.csv contains 'Severity' without variance
# Convert Y_data_variance to a 1D array
Y_data_variance = Y_data_variance.values.ravel()
# Split the data into training and testing sets
X_train, X_test, y_train, y_test = 
train_test_split(X_data, Y_data_variance, test_size=0.2, random_state=42)
# Create and train the Bayesian Ridge Regression model
model = BayesianRidge(
alpha_1=2.0,
alpha_2=0.01,
lambda_1=0.001,
lambda_2=0.01,
alpha_init=None,
lambda_init=None
)
model.fit(X_train, y_train)
\end{lstlisting}

Two key metrics were employed to gauge the model's accuracy: the coefficient of
determination (R2) and the mean-squared-error (MSE). These metrics serve as
fundamental benchmarks in assessing the model's performance. The coefficient of
determination, R2, provides insights into how well the model fits the data or, in simpler terms, the percentage of variance explained by the model. Meanwhile, the MSE
quantifies the overall accuracy of the model by measuring the square of the differences
between predicted values and expected values.

Across ten separate iterations of model training and evaluation, an average MSE
(after adjustment for naturally large index sizes) of 0.1220 and an average R2 value of
0.72 were observed for the precise (no artificial variance) data sets. It's essential to
recognize that while this may be considered a rudimentary and straightforward model,
these results can be deemed as quite successful. The primary source of error in the model
was attributed to the lack of meticulously sorted data, as previously discussed in Section
V of the research. However, this limitation can be reasonably addressed through more
organized and meticulous clinical data collection practices.

\begin{figure}[H]
    \centering
    \includegraphics[width=1\linewidth]{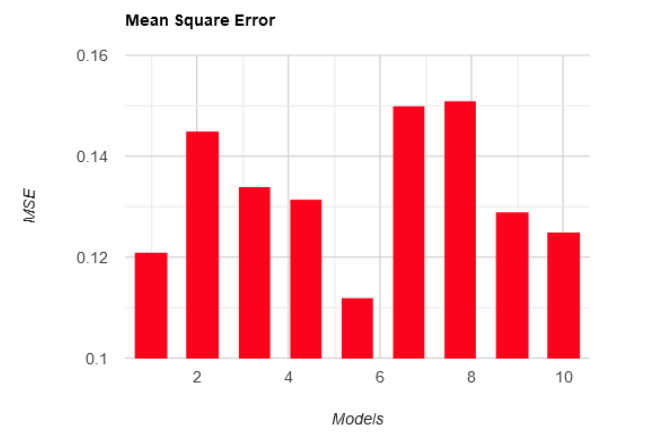}
    \caption{MSE Bar Chart}
    \label{fig:enter-label}
\end{figure}

\begin{figure}[H]
    \centering
    \includegraphics[width=1\linewidth]{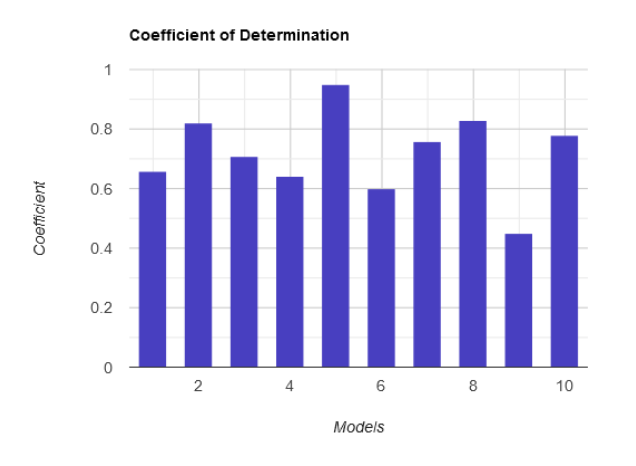}
    \caption{Coefficient of Determination Bar Chart}
    \label{fig:enter-label}
\end{figure}


\begin{lstlisting}[language=Python, caption={Model testing and error collection}, label={lst:python}]
# Test the model on the test dataset for precise data
y_pred = model.predict(X_test)
# Print evaluation metrics for the model's performance 
on the test data with precise values
mse = mean_squared_error(y_test, y_pred)
r2 = r2_score(y_test, y_pred)
print("Mean Squared Error (MSE) on Test Data with Precise Values:", mse)
print("R-squared (R2) on Test Data with Precise Values:", r2)
\end{lstlisting}

Another way to better fit the model to the data would be to tweak the Bayesian
hyper parameters. In the context of Bayesian Ridge Regression and the hyperparameters
scale and rate, it's worth noting that these hyperparameters are often modeled using a
Gamma distribution. In Bayesian Ridge Regression, the scale and rate hyperparameters,
denoted as $\alpha$ (alpha) and $\lambda$ (lambda) respectively, are often assigned prior distributions to
incorporate prior beliefs about their values. A common choice for these prior
distributions is the Gamma distribution.

The Gamma distribution is a probability distribution that is characterized by two
parameters: a shape parameter ($\alpha$) and a rate parameter ($\beta$). The shape parameter ($\alpha$)
determines the shape of the distribution, while the rate parameter ($\beta$) influences the scale
of the distribution. When modeling the scale and rate hyperparameters in Bayesian Ridge
Regression, the Gamma distribution is particularly useful because it allows for a flexible
representation of uncertainty and prior beliefs.

The scale parameter ($\alpha$) in the Gamma distribution corresponds to the shape
parameter in Bayesian Ridge Regression. It influences the shape of the distribution of the
hyperparameter values, capturing the degree of belief in the range and variation of the
hyperparameter.

The rate parameter ($\beta$) in the Gamma distribution corresponds to the rate
parameter ($\lambda$) in Bayesian Ridge Regression. It controls the scale or precision of the
distribution, reflecting how strongly one believes in the specific values of the
hyperparameter.

By choosing a suitable Gamma distribution for the priors on $\alpha$ and $\lambda$, researchers
can express their prior beliefs about the scale and rate hyperparameters in a way that
influences the model's behavior. For instance, a Gamma distribution with a higher shape
parameter $\alpha$ and a lower rate parameter $\beta$ would express a prior belief in larger values for
the hyperparameters, encouraging stronger regularization (for $\alpha$) or less precise
coefficient estimates (for $\lambda$). Conversely, a Gamma distribution with a lower $\alpha$ and a
higher $\beta$ would express a prior belief in smaller values for the hyperparameters, leading
to weaker regularization or more precise coefficient estimates.

It's worth noting that while there is room for improving the accuracy of the model,
it's essential to consider the practical implications of such enhancements. In the real-world context of healthcare, it's unlikely that healthcare workers would be so
stretched that minute differences in severity indexes would significantly impact clinical
decisions. Instead, the severity index serves as a valuable tool for gaining a broad
understanding of a patient's care needs, which can aid in the allocation of healthcare
resources based on priority. For instance, healthcare professionals often use the severity index to categorize
patients into larger priority groups rather than relying on the granular specificity that the
model offers. In this broader view, many of the minor accuracy errors become
inconsequential and go largely unnoticed, as healthcare providers tend to prioritize
patients' needs based on the larger clinical context.

In conclusion, the model's accuracy assessment revealed promising results, with
room for improvement primarily related to data organization. While enhancing accuracy
is feasible through better data practices, it's important to recognize that the practical
utility of the severity index lies in its ability to provide a broad overview of patient care
needs, aligning with healthcare professionals' preference for broader categorizations.
Therefore, while accuracy improvements are beneficial, they must be weighed against the
pragmatic realities of healthcare delivery.

\section{Conclusion}
In conclusion, Bayesian Ridge Regression is a powerful AI model that can be used
to bring cutting edge virus analysis to healthcare professionals around the world.
Implementing AI modeling in data-heavy and detail-oriented tasks is the next step to
improve efficiency in both research and clinical applications.

\section{Future Applications}
Throughout this paper, RSV was used aws the primary disease for counting and
predictions. RSV was chosen as the baseline disease for the severity calculator because it
primarily affects infants, who have few other relevant health complications that increase
the complexity of the data. In future research, more parameters and data stripping can
make models for more complex diseases achievable. In addition to this, having a
compilation of relevant diseases can shift the model’s abilities to diagnosing from only
prognosing.

\section{Works Cited}
“Body Mass Index-for-Age (BMI-for-Age).” World Health Organization, World Health
Organization, www.who.int/toolkits/child-growth-standards/standards/body-mass-index-for-age-
bmi-for-age. Accessed 7 Oct. 2023.

“Comparing Linear Bayesian Regressors.” Scikit,
scikit-learn.org/stable/auto\_examples/linear\_model/plot\_ard.html\#sphx-glr-auto-e
xamples-linear-model-plot-ard-py. Accessed 7 Oct. 2023.

“Curve Fitting with Bayesian Ridge Regression.” Scikit,
scikit-learn.org/stable/auto\_examples/linear\_model/plot\_bayesian
\_ridge\_curvefit.
html\#sphx-glr-auto-examples-linear-model-plot-bayesian-ridge-curvefit-py.
Accessed 7 Oct. 2023.

Hong, Bryan, and Jai Pal. “Improving Viral Diagnostic Methods: A Plasmonic
Nanoparticle Virion Counting and Interpretation System Utilizing MATLAB.”
arXiv.Org, 10 Sept. 2023, arxiv.org/abs/2309.04924.

“Sklearn.Linear\_model.Bayesianridge.” Scikit,
scikit-learn.org/stable/modules/generated/sklearn.linear\
\_model.BayesianRidge.ht
ml. Accessed 7 Oct. 2023.

\end{document}